\documentclass{sig-alt-release2}
\usepackage{multirow} 

\begin{document}
%
\conferenceinfo{GECCO'09,} {July 8--12, 2009, Montr\'eal, Qu\'ebec, Canada.} 

\CopyrightYear{2009}

\crdata{978-1-60558-505-5/09/07} 

\title{Improving NSGA-II with an Adaptive Mutation Operator}

\numberofauthors{2} 

\author{
\alignauthor
Arthur G. Carvalho \\
       \affaddr{School of Computer Science}\\
       \affaddr{University of Waterloo}\\
       \affaddr{Waterloo, Ontario, Canada}\\
       \email{a3carval@cs.uwaterloo.ca}
\alignauthor
Aluizio F. R. Araujo \\
       \affaddr{Informatics Center}\\
       \affaddr{Federal University of Pernambuco}\\
       \affaddr{Recife, Pernambuco, Brazil}\\
       \email{aluizioa@cin.ufpe.br}
}

\maketitle
\begin{abstract}
The performance of a Multiobjective Evolutionary Algorithm (MOEA) is crucially dependent on the parameter setting of the operators. The most desired control of such parameters presents the characteristic of adaptiveness, i.e., the capacity of changing the value of the parameter, in distinct stages of the evolutionary process, using feedbacks from the search for determining the direction and/or magnitude of changing. Given the great popularity of the algorithm NSGA-II, the objective of this research is to create adaptive controls for each parameter existing in this MOEA. With these controls, we expect to improve even more the performance of the algorithm.

In this work, we propose an adaptive mutation operator that has an adaptive control which uses information about the diversity of candidate solutions for controlling the magnitude of the mutation. A number of experiments considering different problems suggest that this mutation operator improves the ability of the NSGA-II for reaching the Pareto optimal Front and for getting a better diversity among the final solutions.
\end{abstract}




\section{Introduction}
The term optimization, in the field of mathematics, refers to the study of problems in which we are looking for optimal solutions, minimum or maximum, for a given function. These solutions are obtained through systematic changes in the values of the variables. When we want to optimize systematically and simultaneously various objective functions (usually conflicting between themselves), we will have the process known as multiobjective optimization.

A good algorithm created for solving multiobjective optimization problems must: 1) find multiple Pareto optimal solutions and 2) find a good diversity of solutions on the obtained Pareto front (close to an uniform distribution)\cite{deb:EmoBook}. 

Variations of evolutionary algorithms, known as Multiobjective Evolutionary Algorithms (MOEAs), are the metaheuristic best known for solving multiobjective optimization problems. Due to the characteristics inherited from the evolutionary computing, these algorithms have operators with parameters that need to be configured. Moreover, the performance of a MOEA is crucially dependent on the parameter setting of these operators.

The most desired control of such parameters presents the characteristic of adaptiveness, i.e., the capacity of changing the value of the parameter, in distinct stages of the evolutionary process, using feedbacks from the search for determining the direction and/or magnitude of changing. However, MOEAs usually employs stochastic operators with static parameters. 

According to Eiben and Smith \cite{eiben:book} a run of an evolutionary algorithm is a process intrinsically dynamic and adaptive. Then, this static approach can result in an inefficient convergence to the Pareto optimal solutions and a failure for creating an (almost) uniform distribution of final solutions on the obtained Pareto front.

Given the great popularity of the algorithm \textit{Non-dominated Sorting Genetic Algorithm II} (NSGA-II) \cite{deb:NSGA-II}, we propose to create adaptive controls for each parameter existing in this MOEA for increasing even more its ability for reaching the Pareto optimal front and for getting a better diversity among the final solutions.

Within the context presented, we propose in this work an adaptive mutation operator which uses information about the diversity of candidate solutions for controlling the magnitude of the mutation. 

The rest of this paper is organized as follows. Section 2 presents the concept of \textit{crowding distance} \cite{deb:NSGA-II}, a density estimator that will provide information for controlling the magnitude of the mutation. Section 3 describes the adaptive mutation operator proposed. The experiments and the statistical validation of the results are described in Section 4. Finally, Section 5 summarizes the results of this work and proposes additional topics for further research.

\section{Crowding Distance}
The crowding distance is an important concept proposed by Deb \textit{et. al.} \cite{deb:NSGA-II} in his algorithm NSGA-II. It serves for getting an estimate of the density of solutions surrounding a particular solution in the population. More specifically, the crowding distance for a point $i$ (called $i_{distance}$) is an estimate of the size of the largest cuboid enclosing $i$ without including any other point in the population. It is calculated by taking the average distance of the two points on either side of $i$ along each of the objectives. The algorithm used for calculating the crowding distance for each point in a population $L$ is:

\begin{tabbing}
crow\=ding-distance-assignment($L$):\\
    \> $l = |L|$\\
    \> for \=\+each $i\in L$\\
    \>     $L[i]_{distance} = 0$ \- \\
    \> for \=\+each objective $m$ \+ \\
           $L=sort(L,m)$ \\
           $L[1]_{distance}=L[l]_{distance} = \infty $ \\
           for \= $i=2$ to $(l-1)$ \+ \\
               $L[i]_{distance} \mbox{+=}$ \= $\underline{f_m(i+1) - f_m(i-1)}$ \+ \\
$\quad \max{f_{m}} - \min {f_m} $ 
\end{tabbing}

In the first line, it is assigned the size of the population $L$ to the variable $l$. Following this operation, there is a loop responsible for initializing with $0$ the $i_{distance}$ of each element $i$ of the population $L$.

In the fourth line, each objective $m$ is selected at a time and the population is sorted in a ranking according to the value for $m$. The $i_{distance}$ value for solutions in the first and in the last position is assigned as infinity ($\infty$) for preserving solutions with extreme values.

The inner loop presents in the seventh line updates the $i_{distance}$ value for each remaining solution \textit{i} from position $2$ to $l-1$. First, it is calculated the $m$-th objective function value for the neighbors of \textit{i}. Thereafter, it is calculated the difference between the highest and the lowest value. Finally, the $i_{distance}$ value for $i$ is updated by the sum of its previous value with the normalized result of that subtraction. Figure 1 shows an illustration of this calculation for a given solution $i$. In this scenario, the $i_{distance}$ value for $i$ will be $r+s$ where:

\begin{eqnarray*}
s=\frac{f_1(i-1)-f_1(i+1)}{f_1(a)-f_1(z)} \quad \mbox{and} \quad r=\frac{f_2(i+1)-f_2(i-1)}{f_2(z)-f_2(a)}
\end{eqnarray*}

\begin{figure}[!htbp]
  \begin{center}
    \includegraphics[scale=1]{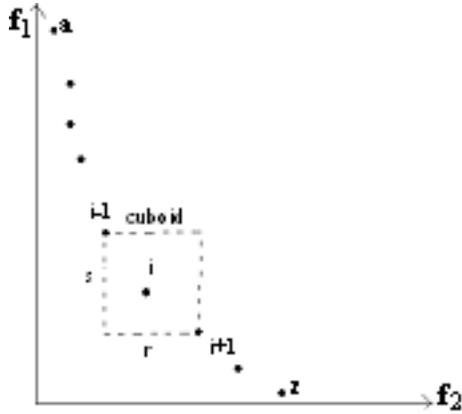}
    \caption{The crowding distance calculation.}
    \label{fig:1}
  \end{center}
\end{figure}

\section{Adaptive Mutation Operator}

According to Eiben and Smith \cite{eiben:book}, an adaptive parameter control uses feedback from the search for serving as input to a mechanism used for determining the direction and/or magnitude of changing. Using the well known static mutation operator proposed by Deb and Goyal \cite{deb:geneAS} together with an adaptive parameter control for updating its parameter, this section presents the adaptive mutation operator created for improving even more the performance of the algorithm NSGA-II. 

In the original (static) version of the mutation operator, the current value of a continuous variable is changed to a neighboring value using a polynomial probability distribution. This distribution has its mean at the current value of the variable and its variance as a function of a parameter $n$. This parameter will define the strength of the mutation and we are interested in adaptively changing its value.

Besides this parameter, the polynomial probability distribution depends on a factor of disturbance $\delta$ for calculating the mutated value as can be seen in the following equation:

\begin{equation}
P(\delta) = 0.5(n+1)(1-|\delta|)^n
\end{equation}
where $\delta \in [-1,1]$. Figure 2 shows this distribution for some values of $n$.

\begin{figure}[!htbp]
  \begin{center}
    \includegraphics[scale=0.55]{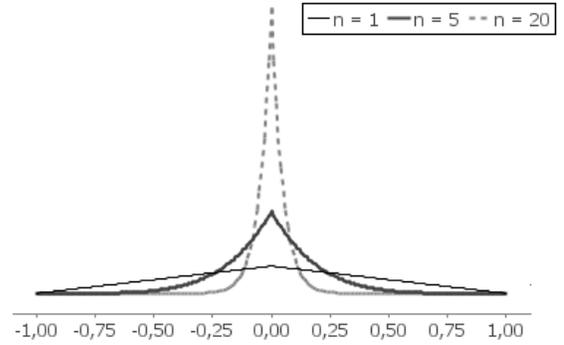}
    \caption{Probability distribution for creating a mutated value.}
    \label{fig:2}
  \end{center}
\end{figure}

Initially, for creating a mutated value we need to generate a random number $u \in [0,1]$. Thereafter, the equation 2 (obtained from equation 1) can be used for calculating the factor of disturbance $\overline{\delta}$ corresponding to $u$:

\begin{equation}
\overline{\delta} = \left\{
\begin{array}{ll}
(2u)^\frac{1}{n+1} - 1 & \mbox{if } u < 0.5\\
 1-[2(1-u)]^\frac{1}{n+1} & \mbox{if } u \ge 0.5\\
\end{array}
\right.
\end{equation}

In the end, the mutated valued is calculated using the following equation:

\begin{equation}
c=p+\overline{\delta}\Delta_{max}
\end{equation}
where $c$ is the mutated value, $p$ is the original value and $\Delta_{max}$ is the maximum disturbance allowed in the value of $p$ (it was defined here as the difference between the maximum and the minimum value for the decision variable).

To change the variance of the probability distribution (the parameter $n$ in equation 1) in an adaptive way, we will use two empirical facts observed. First, the initial solutions are dispersed in the search space and distant from the Pareto optimal front. Furthermore, the difference between the greatest $i_{distance}$ value not infinite and the lowest $i_{distance}$ value is lifted. In this scenario, it is necessary to apply a strong mutation for ensuring a quicker convergence to the Pareto optimal Front and a fast attainment of distinct solutions.

Second, at the end of the evolutionary process it will be expected solutions closer to the Pareto optimal front due to the efficacy of the NSGA-II. Moreover, the difference between the greatest (not infinite) and the lowest $i_{distance}$ value is reduced. Now, it is necessary to apply a soft mutation for avoiding destroying solutions previously generated and for trying to approximate them to the Pareto optimal front.

So, the main ideas exploited by the adaptive control are to use information about the difference between the greatest (not infinite) and the lowest $i_{distance}$ value and about the current stage of the evolutionary process. Due to the fact that the NSGA-II calculates the  $i_{distance}$ for all individuals in the current population before applying evolutionary operators, it will not be necessary to re-calculated it again. So, we just have to calculate $\Delta$, the difference between the greatest (different of $\infty$) and the lowest $i_{distance}$ value:

\begin{eqnarray*}
\Delta &=& \max_{1 \leq i \leq |L|} g(i_{distance}) - \min_{1 \leq i \leq |L|}i_{distance} \\
\mbox{where:} \quad g(x) &=& 0 \quad if \quad x = \infty \\
&=& x \quad otherwise
\end{eqnarray*}

The next step is to use information about the current generation $t$ of the evolutionary process. For ensuring that it will have an acceptable weight in the update of the parameter, we applied on it a logistic function. So, the second step taken by the controller is to calculate the function:

\begin{equation}
sigm(t) = \frac{1}{1 + e^{-0.07t}}
\end{equation}
where $t$ is the current generation. The inspiration for using such function is the fact that it would fit perfectly into our proposal because we would like to apply a strong mutation in the early stages of the evolutionary process and gradually reduce its value during the process. The constant value $-0.07$ is used because the value of $e^{-0.07t}$ will be approximately $0$ when $t \rightarrow \infty$. Actually, when $t$ is greater than $100$, the function $sigm(t)$ will practically stop influencing the mutation because its value will be equals to $1$.

It is useful to cite that the new value for the parameter $n$ has to be inversely proportional to $\Delta$. This happens due to the fact that for higher values of $\Delta$ it will be necessary to apply a strong mutation and, consequently, it will be needed a lower value for $n$ to increase the variance of the probability distribution. Furthermore, the new value for $n$ has to be directly proportional to the $sigm(t)$ due to the fact that for higher values of $t$ it will be needed a soft mutation and, consequently, it will be needed a higher value for $n$ to reduce the variance of the probability distribution. In the end, the last step taken by the controller is to update $n$, before applying a mutation in the current generation, as follows:

\begin{equation}
n=\frac{sigm(t)}{\Delta}
\end{equation}

\section{Experiments}

In order for evaluating the performance of the proposed adaptive mutation operator, this section provides a comparative study among different settings for the NSGA-II. The first one uses the original mutation operator proposed by Deb and Goyal \cite{deb:geneAS} with $n = 5$ (for representing a strong mutation). The second configuration also uses this mutation operator, but this time with $n = 20$ (for representing a smooth mutation). At least, the third configuration is represented by the adaptive mutation operator proposed here.

The remaining parameters are the same for all settings. We used a population size of $20$ individuals (this small value was chosen for making the mutation more valorous), a crossover probability of $0.9$, a mutation probability of $1/n$ (where $n$ is the number of variables). The variables were treated as real numbers and the simulated binary crossover operator (SBX) \cite{deb:geneAS} was used. For all experiments, the implementation used as reference was proposed by Durillo \textit{et al} \cite{jMetal}.

The problems used in experiments were chosen based on characteristics usually present in real problems \cite{Coello:Book}: continuous Pareto optimal front vs. discontinuous Pareto optimal front; convex Pareto optimal front vs. non-convex Pareto optimal front; uniformly represented Pareto optimal Front vs. non-uniformly represented Pareto optimal front.

The first problem used was proposed by Fonseca and Fleming \cite{Fonseca} (called here as FON2). The next four problems used (ZDT1, ZDT2, ZDT3, ZDT6) were proposed by Zitzler \textit{et al} \cite{ComparisonMOPs} and belong to a test suite called ZDT.

Due to the fact that the convergence to the Pareto optimal front and the maintenance of a diverse set of solutions are two different goals of the multiobjective optimization, it will be need two different metrics for deciding the performance of a setting in an absolute manner \cite{deb:EmoBook}.

The first metric used, called Generational Distance (GD) \cite{MOPs}, is responsible for finding the closeness of the obtained set of solutions to the Pareto optimal front as follows:

\begin{equation}
GD = \frac{(\sum_{i=1}^{|Q|}d_i^2)^\frac{1}{2}}{|Q|}
\end{equation}
where $Q$ is the set of the obtained solutions and $d_i$ is the Euclidean distance between the solution $i \in Q$ and the nearest member of the Pareto optimal front as exhibited below:

\begin{equation}
d_i = \min_{k=1}^{|P*|} \sqrt{\sum_{m=1}^{M}\left( f_m^{(i)} - f_m^{*(k)} \right)^2}
\end{equation}
where $P*$ is the Pareto optimal front and $f_{m}^{*(k)}$ is the \textit{m}-th objective function value of the  \textit{k}-th member of $P*$. This metric has the constraint that it is necessary the Pareto optimal front. Here, for each problem utilized in the experiments we used the front provided by Coello \textit{et al} \cite{Coello:Book}. It is useful to note that before calculating this distance measure, it is necessary to normalize the objective function values.

The second metric used measures the spread of the obtained set of solutions calculating the non-uniformity in the distribution. It was proposed by Deb \textit{et al} \cite{deb:NSGA-II} as follows:

\begin{equation}
\Delta = \frac{d_f + d_l + \sum_{i=1}^{|Q|-1}|d_i - \bar{d}|}{d_f + d_l + (|Q|-1)\bar{d}}
\end{equation}
where $d_i$ is any distance metric between neighboring solutions, $\bar{d}$ is the mean value of these distance measures and $d_f$ and $d_l$ are the distances between boundary solutions from the set of obtained solutions and the Pareto optimal front. For both metrics, a lower value implies in a better result.

In the end, we run each configuration $100$ independent times until the $100$-th generation in each problem. The obtained results according to the metrics spread and generational distance are shown respectively in Table 1 and Table 2. In each row of these tables, we have the upper cell containing the mean for the $100$ runs (the lower value is highlighted with bold font) and below it a cell containing the standard deviation. Moreover, for the rows representing the settings $n = 5$ and $n = 20$ we have a bottom cell containing the results of the use of the statistical test called \textit{test t} with a confidence level of $95\%$.

This test is applicable for comparing two samples of two populations normally distributed, not necessarily of the same size, where the mean and the variance of the population are unknown. We used this test for understanding whether there is a statistically significant difference between the results produced by the setting $n = 5$ or $n = 20$ and the results obtained by the adaptive approach. The value "$+$" indicates that the adaptive approach will have a lower value with $95\%$ of confidence, the value "$-$" represents that the adaptive approach will have a higher value with $95\%$ of confidence and the value "$\approx$" means that there is not statistically significant difference between the approaches. 

\begin{table}[!htb]
  \label{table1}
  \begin{center}
    \caption{Results obtained by the metric spread}
    \vspace{0.2cm}
    \footnotesize
    \begin{tabular}{|c|c|c|c|c|c|}\hline
      Setting & ZDT1 & ZDT2 & ZDT3 & ZDT6 & FON2\\ \hline
     \multirow {3}{*}{n = 5} & 0.443 & 0.619 & 0.574 & 0.482 & 0.441  \\  \cline{2-6}
                             & 0.078 & 0.175 & 0.062 & 0.201 & 0.086  \\ \cline{2-6}
                             & $\approx$ & + & + & $\approx$ & +      \\ \hline
     \multirow {3}{*}{n = 20}& 0.609 & 0.940 & 0.677 & 0.646 & 0.412  \\  \cline{2-6}
                             & 0.070 & 0.076 & 0.063 & 0.213 & 0.083  \\ \cline{2-6}
                             & + & + & + & + & $\approx$  \\ \hline
\multirow {2}{*}{adaptive} &  \textbf{0.428} &  \textbf{0.463} & \textbf{0.561} & \textbf{0.462} & \textbf{0.410}  \\  \cline{2-6}
                             & 0.065 & 0.077 & 0.051 & 0.140 & 0.091  \\ 
\hline\end{tabular}
  \end{center}
\end{table}

\begin{table}[!htb]
  \label{table2}
  \begin{center}
    \caption{Results obtained by the metric GD}
    \vspace{0.2cm}
    \footnotesize
    \begin{tabular}{|c|c|c|c|c|c|}\hline
      Setting & ZDT1 & ZDT2 & ZDT3 & ZDT6 & FON2\\ \hline
     \multirow {3}{*}{n = 5} & 0.012 & 0.011 & 0.008 & 0.016 & \textbf{0.005}  \\  \cline{2-6}
                             & 0.008 & 0.007 & 0.004 & 0.040 & 0.001  \\ \cline{2-6}
                             & + & + & + & + & $\approx$     \\ \hline
     \multirow {3}{*}{n = 20}& 0.077 & 0.212 & 0.052 & 0.039 & \textbf{0.005}  \\  \cline{2-6}
                             & 0.023 & 0.159 & 0.016 & 0.052 & 0.001  \\ \cline{2-6}
                             & + & + & + & + & $\approx$  \\ \hline
\multirow {2}{*}{adaptive} &  \textbf{0.008} &  \textbf{0.005} & \textbf{0.006} & \textbf{0.008} & \textbf{0.005}  \\  \cline{2-6}
                             & 0.005 & 0.002 & 0.003 & 0.019 & 0.001  \\ 
\hline\end{tabular}
  \end{center}
\end{table}

As can be seen from the tables, the adaptive mutation operator got the lowest means for both metrics in all problems. Furthermore, in 3 of 5 problems the adaptive approach  obtained the lowest standard deviation for the metric spread and in all problems it got the lowest standard deviation for the metric generational distance. 

Looking for the results of the \textit{test t}, the adaptive approach was superior to the setting with $n = 5$ in 3 problems for the metric spread and in 4 problems for the metric generational distance. In relation to the setting $n = 20$, the adaptive approach was better in 4 problems for both metrics.

\section{Conclusions}

This paper presented the first step for creating adaptive controls for each parameter present in the algorithm NSGA-II to improve even more its performance. We proposed an adaptive mutation operator that uses information about the diversity of the population, through the concept of crowding distance, for controlling the strength of the mutation.

Running the algorithm NSGA-II on five different problems, we compared the results obtained by the adaptive approach with the results obtained by two different static settings: a setting that applied a strong mutation and a setting that applied a smooth mutation. The experimental results have shown that the approach proposed outperformed both settings in convergence to the Pareto optimal Front and in diversity of the final solutions. A statistical test was done to prove the relevance of the results.

While the approach seems interesting, it is clear that more work will be necessary to understand its impact on the search. Moreover, a clear empirical study is required to demonstrate its significance. It is useful to cite that this approach can also be used for controlling parameters of other operators. For instance, the parameter that controls the proximity of the offspring from the parents in the crossover operator (SBX) proposed by Deb \cite{deb:geneAS} can be controlled in such way that new solutions staying closer of parents with higher crowding distance. This would help in increasing diversity.

%

%

\begin{thebibliography}{1}

\bibitem{Coello:Book}
C.~A.~C. Coello, G.~B. Lamont, and D.~A.~V. Veldhuizen.
\newblock {\em Evolutionary Algorithms for Solving Multi-Objective Problems}.
\newblock Springer, 2007.

\bibitem{deb:EmoBook}
K.~Deb.
\newblock {\em Multi-Objective Optimization using Evolutionary Algorithms}.
\newblock John Wiley and Sons, 2001.

\bibitem{deb:geneAS}
K.~Deb and M.~Goyal.
\newblock A combined genetic adaptive search (geneas) for engineering design.
\newblock {\em Computer Science and Informatics}, 26(4):30--45, 1996.

\bibitem{deb:NSGA-II}
K.~Deb, A.~Pratap, S.~Agarwal, and T.~Meyarivan.
\newblock A fast and elitist multiobjective genetic algorithm: Nsga-ii.
\newblock {\em IEEE Transactions on Evolutionary Computation}, 6(2):182--197,
  April 2002.

\bibitem{jMetal}
J.~J. Durillo, A.~J. Nebro, F.~Luna, B.~Dorronsoro, and E.~Alba.
\newblock {{jMetal}: A Java Framework for Developing Multi-Objective
  Optimization Metaheuristics}.
\newblock Technical Report ITI-2006-10, Departamento de Lenguajes y Ciencias de
  la Computaci\'on, University of M\'alaga, E.T.S.I. Inform\'atica, Campus de
  Teatinos, December 2006.

\bibitem{eiben:book}
A.~E. Eiben and J.~E. Smith.
\newblock {\em Introduction to evolutionary computing}.
\newblock Springer, 2003.

\bibitem{Fonseca}
C.~M. Fonseca and P.~J. Fleming.
\newblock Multiobjective genetic algorithms made easy: Selection, sharing and
  mating restriction.
\newblock In {\em Proceedings of the First International Conference on Genetic
  Algorithms in Engineering Systems: Innovations and Applications}, pages
  45--52, 1995.

\bibitem{MOPs}
D.~A.~V. Veldhuizen.
\newblock {\em Multiobjective evolutionary algorithms: classifications,
  analyses, and new innovations}.
\newblock PhD thesis, Air Force Institute of Technology, Wright Patterson AFB,
  OH, USA, 1999.

\bibitem{ComparisonMOPs}
E.~Zitzler, K.~Deb, and L.~Thiele.
\newblock Comparison of multiobjective evolutionary algorithms: Empirical
  results.
\newblock {\em Evolutionary Computation}, 8(2):173--195, 2000.

\end{thebibliography}
\end{document}